\documentclass[letterpaper, 10 pt, conference]{ieeeconf}  % Comment this line out if you need a4paper

\IEEEoverridecommandlockouts                              % This command is only needed if 
\usepackage{amssymb}  % assumes amsmath package installed
\usepackage{amsmath}
\usepackage{cite}
 
\usepackage{pdfpages}
\usepackage{graphics}
\usepackage{lipsum}
\usepackage{mathtools}
\usepackage{cuted}
\usepackage[T1]{fontenc}
\usepackage[utf8]{luainputenc}
\usepackage{amsbsy}
\usepackage{algorithm}
\usepackage{algpseudocode}
\usepackage{algpascal}
\usepackage{bm}
\usepackage{hyperref}
\usepackage[font=small,skip=0pt]{caption}
\usepackage{cite}
\usepackage{epsfig}
\title{\LARGE \bf
Distal-Stable Beam for Continuum Robots
}

\author{Ryouichi Saito$^1$, Takahiro Koide$^1$, Yuya Tanaka$^2$, Yasutaka Nakashima$^1$, Motoji Yamamoto$^1$, Ayato Kanada$^2$% <-this % stops a space 
\thanks{$^1$Ryouichi Saito, Takahiro Koide, Yasutaka Nakashima, Motoji Yamamoto are with the Department of Mechanical Engineering at Kyushu University, 744 Motooka, Nishi-ku, Fukuoka, 819-0395, Japan.}% <-this % stops a space
\thanks{$^2$Takahiro Koide, Ayato Kanada are with the Department of Graduate School of Informatics and Engineering at The University of Electro-Communications, 1-5-1 Chofugaoka, Chofu-shi, Tokyo, 182-8585
        {\tt\small}}%%
}

\begin{document}

\maketitle

%%%%%%%%%%%%%%%%%%%%%%%%%%%%%%%%%%%%%%%%%%%%%%%%%%%%%%%%%%%%%%%%%%%%%%%%%%%%%%%%
\begin{abstract}
Continuum robots are well suited for constrained environments but suffer from low distal stiffness, resulting in large posture errors under external loads. In this paper, we propose a novel structural primitive, the Distal-Stable Beam, which achieves a strong stiffness gradient through purely geometric design, maintaining compliance in the intermediate section while ensuring high distal rigidity. The structure consists of two parallel rods and one convergent rod constrained by guide disks, introducing geometric coupling that suppresses deformation modes and preserves distal posture. Experiments show that the distal stiffness is 12 times higher than at the center, corresponding to an approximately 100-fold improvement over a conventional cantilever beam. The proposed mechanism enables simultaneous compliance and distal stability without active stiffness modulation, providing a new design approach for continuum robots requiring both safety and precision.
\end{abstract}

%%%%%%%%%%%%%%%%%%%%%%%%%%%%%%%%%%%%%%%%%%%%%%%%%%%%%%%%%%%%%%%%%%%%%%%%%%%%%%%%
\section{INTRODUCTION}

\begin{figure}[t!]
\centering
\includegraphics[width=0.95\linewidth]{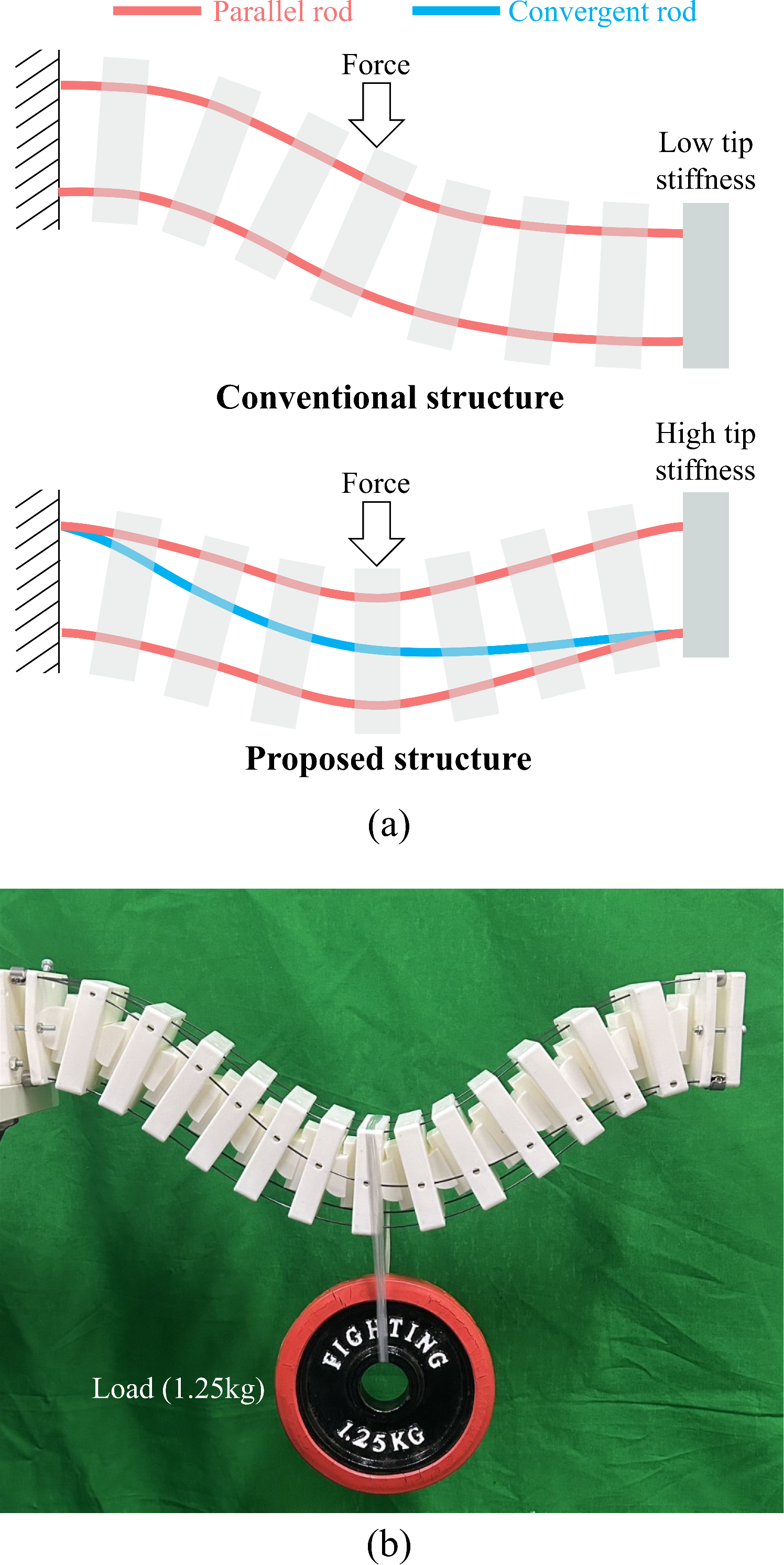}
\caption{Concept of the distal-stable beam, illustrating simultaneous compliance and high distal stiffness.}
\label{concept}
\end{figure}

Continuum robots have emerged as a promising solution for navigating complex and constrained environments due to their inherent compliance and continuous bending capabilities~\cite{da2020challenges,singh2014continuum,kim2013soft,trivedi2008soft}. Their ability to adapt their shape makes them well suited for applications ranging from minimally invasive surgery to delicate manipulation in cluttered environments. However, this inherent softness introduces a fundamental trade-off: significant distal posture errors often occur when the robot interacts with external obstacles or carries payloads~\cite{FAN2024102811}. This lack of distal precision under load limits their use in tasks requiring both high maneuverability and stable distal posture (Fig.~\ref{concept}(a)).

To address the issue of low distal stiffness, variable stiffness mechanisms have been extensively investigated, employing techniques such as jamming mechanisms~\cite{wang2024soft,cheng2012design,kim2013novel} or low-melting-point materials~\cite{zhang2023bioinspired,wei2021analysis,wang2022flexible}. These approaches enable the robot to transition from a compliant state to a rigid one when needed. However, such methods typically increase the stiffness of the entire structure. This global stiffening generally requires larger actuation forces and increased energy consumption, and it may introduce a non-negligible delay during the stiffness transition. 
This delay arises from internal state changes, such as pressure redistribution in jamming systems and thermal phase transitions in low-melting-point materials. While these effects can be mitigated through engineering optimization, they cannot be entirely eliminated.
Such global stiffening and associated delays may compromise the fundamental advantage of continuum robots, namely their ability to maintain passive compliance during sudden impacts.

Another approach to enhancing stiffness involves optimizing the physical arrangement of actuation tendons or rods. The low stiffness of continuum robots primarily originates from the presence of passive deformation modes. Most existing continuum robots employ parallel actuation paths, where tendons or rods are routed parallel to the backbone centerline. When a lateral load is applied to such a structure, it typically exhibits an S-shaped deformation mode~\cite{6661461}. Because this deformation can occur even when the lengths of the actuation tendons remain unchanged, the resulting tip stiffness remains inherently low, leading to poor positioning accuracy.

Recent studies have suggested that combining parallel actuation paths with convergent actuation paths can suppress these S-shaped deformation modes, thereby improving tip stiffness~\cite{10.3389/frobt.2021.629871,8606257}. While this structural strategy is effective for increasing rigidity, a critical challenge remains: achieving a balance between the overall softness required for safe interaction and the high stiffness required at the distal end. To the best of our knowledge, no prior work has demonstrated a purely geometric mechanism that simultaneously preserves intermediate compliance while maintaining a stable distal posture.

In this paper, we propose the distal-stable beam, a structural primitive that generates a pronounced stiffness gradient along its length (Fig.~\ref{concept}(a) and (b)). The distal tip is highly rigid for payload support and precise manipulation, while the intermediate section remains compliant to enhance safety during unexpected interactions. Even under large intermediate deformations caused by collisions or obstacle contact, the distal tip maintains an almost constant posture.

This behavior is reminiscent of the Fin Ray effect used in compliant grippers for adaptive shape conformity~\cite{ali2019biologically,crooks2016fin,pfaff2011application}. In such structures, deformation allows the gripper to conform to objects while largely preserving the tip position, but without constraining rotational motion. In contrast, the proposed structure maintains both the position and orientation of the distal tip despite large intermediate deformations. Moreover, unlike Fin Ray-based designs with low aspect ratios, the proposed mechanism operates as a slender beam-like structure.

This property originates from a geometric coupling among flexible elements. The structure consists of a simple triad of two parallel rods and one convergent rod, threaded through rigid spacer disks with fixed-pitch apertures. These disks constrain the relative spacing between the rods at each cross-section, effectively coupling their local curvatures. In a cantilever configuration, the beam remains highly compliant and can deform around obstacles, while the geometric coupling ensures minimal deviation of the distal posture. By modeling the rods as coupled geometric curves, we show that the integrated curvature along the beam inherently preserves the distal posture.

The primary contributions of this work are summarized as follows:

\begin{itemize}
\item A theoretical derivation of the distal stability property, modeling the bundled rods as a manifold of geometric curves and identifying the constraints required to preserve distal posture.
\item The development of a passive structural prototype that exhibits high distal rigidity despite its compliant intermediate section, maintaining distal posture under external disturbances.
\item The implementation of an actuated continuum robot based on the proposed rod-bundle architecture, enabling steerable and extensible motion while intrinsically preserving its distal posture through geometric constraints.
\end{itemize}

The remainder of this paper is organized as follows. Section II presents the geometric model and the derivation of invariant properties underlying distal posture preservation. Section III provides numerical validation of the proposed model. Section IV describes the fabrication of the prototype and experimental validation of its kinematic and stiffness properties. Section V demonstrates the application of the proposed principle to an actuated continuum robotic system and evaluates its performance. Finally, Section VI concludes the paper and discusses future directions.

\section{GEOMETRIC MODEL AND INVARIANT ANALYSIS}

\begin{figure}[t!]
\centering
\includegraphics[width=0.95\linewidth]{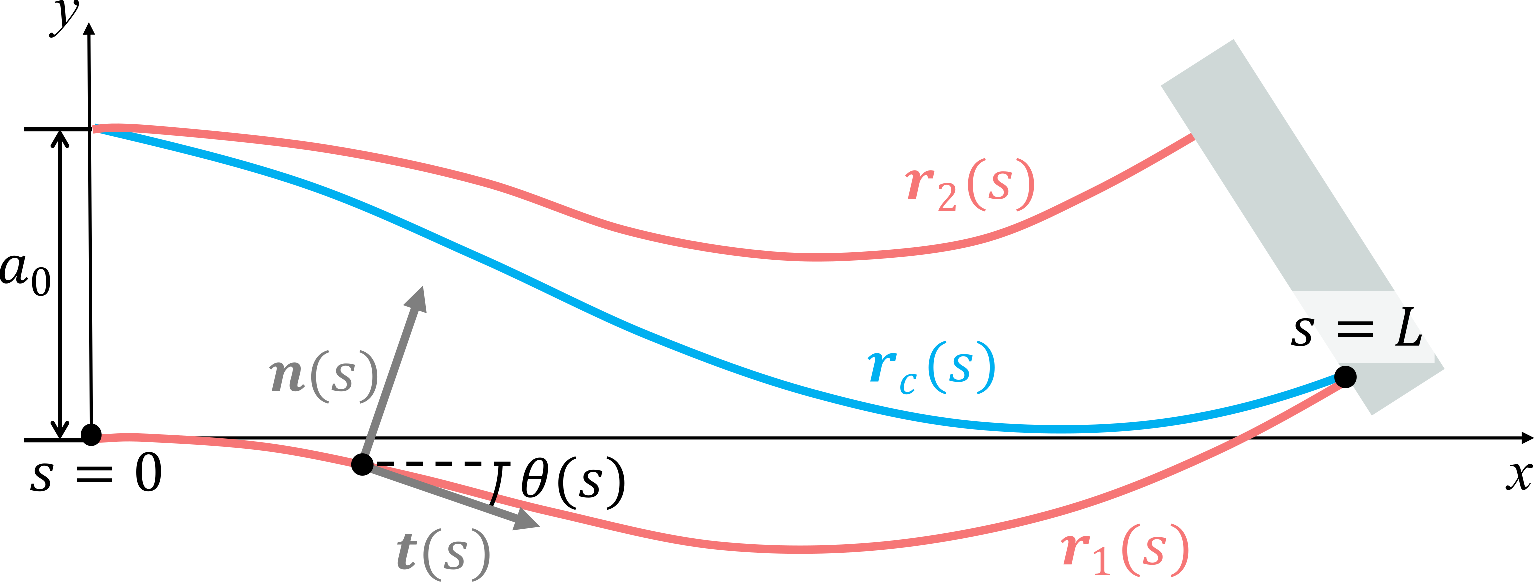}
\caption{Geometric model of the distal-stable beam.}
\label{graph}
\end{figure}

The proposed structure consists of two parallel rods and one convergent rod. Each rod is inextensible, and their total lengths remain constant during deformation. Furthermore, the distance between the rods is structurally constrained. A key property of this structure is that, even when the intermediate section undergoes large deformation, the distal posture almost remains constant. In this section, we mathematically show that this property arises from invariants derived from the length constraints of each rod.

Specifically, we first model each rod as a geometric offset curve on a plane. Next, from the constraint of constant rod length, we derive geometric quantities that remain invariant under deformation. Using these invariants, we then elucidate the mechanism that preserves the distal posture, independent of the deformation of the intermediate section.

\subsection{Geometric Model}
The geometric relationship of the three rods constituting the structure is shown in Fig.~\ref{graph}. In this model, the shape of each rod is defined using a reference arc-length parameter $s \in [0, L]$ as follows:

\noindent \textbf{First Parallel Rod $\bm{r}_1(s)$:} The reference curve of the structure.
\begin{equation}
\bm{r}_1(s) = [x(s), y(s)]^T
\end{equation}
The tangent angle at each point on $\bm{r}_1(s)$ is $\theta(s)$, with the unit tangent vector $\bm{t}(s)$, unit normal vector $\bm{n}(s)$, and curvature defined as $\kappa(s) = d\theta/ds$. The boundary conditions are $\bm{r}_1(0)=(0,0)$ and $\theta(0)=0$.

\noindent \textbf{Second Parallel Rod $\bm{r}_2(s)$:} A rod that maintains a constant normal distance $a_0 \ (a_0>0)$ from the first parallel rod.
\begin{equation}
\bm{r}_2(s) = \bm{r}_1(s) + a_0 \bm{n}(s)
\end{equation}

\noindent \textbf{Convergent Rod $\bm{r}_c(s)$:} A rod whose normal distance from the first parallel rod decreases linearly from $a_0$ at the base to $0$ at the tip.
\begin{equation}
\bm{r}_c(s) = \bm{r}_1(s) + a_0 \left(1 - \frac{s}{L}\right) \bm{n}(s)
\end{equation}

\subsection{Invariant Derivation from Length Constraints}
We define $L_1, L_2$, and $L_c$ as the total lengths of the first parallel rod $\bm{r}_1(s)$, the second parallel rod $\bm{r}_2(s)$, and the convergent rod $\bm{r}_c(s)$, respectively, over the interval $s \in [0, L]$. Since $\bm{r}_1(s)$ is parameterized by the arc length $s$, its total length is inherently $L_1 = L$. Assuming the lengths $L_1, L_2$, and $L_c$ remain constant, we derive two crucial invariants that govern the system.

\vspace{\baselineskip}
\noindent Invariant 1: Fixing the tip angle

We calculate the length $L_2$ of the second parallel rod. The derivative of the offset curve $\bm{r}_2(s)$ with respect to the arc length $s$ is expressed using the Frenet-Serret formulas as follows:
\begin{equation}
\frac{d\bm{r}_2(s)}{ds} = \frac{d\bm{r}_1(s)}{ds} + a_0 \frac{d\bm{n}(s)}{ds} = (1 - a_0 \kappa(s))\bm{t}(s)
\end{equation}
Under the condition that the offset curve does not locally self-intersect ($a_0\kappa(s)<1$), $L_2$ is given by:
\begin{equation}
L_2 = \int_0^L (1 - a_0 \kappa(s)) ds = L - a_0 (\theta(L) - \theta(0))
\end{equation}
Solving for the tip angle $\theta(L)$ with $\theta(0)=0$ yields:
\begin{equation}
\theta(L) = \frac{L - L_2}{a_0}
\end{equation}
Since $L, L_2, a_0$ are all constants, $\theta(L)$ is invariant. Thus, the constraint of the parallel rod fixes the tip angle regardless of the intermediate shape.

\vspace{\baselineskip}
\noindent Invariant 2: Fixing the average angle

The length $L_c$ of the convergent rod $\bm{r}_c(s)$ is calculated. Its derivative with respect to arc length is expanded as:
\begin{align}
\frac{d\bm{r}_c(s)}{ds} &= \frac{d\bm{r}_1(s)}{ds} - \frac{a_0}{L}\bm{n}(s) + a_0\left(1-\frac{s}{L}\right)\frac{d\bm{n}(s)}{ds} \nonumber \\
&= \left( 1 - a_0\left(1-\frac{s}{L}\right)\kappa(s) \right)\bm{t}(s) - \frac{a_0}{L}\bm{n}(s)
\end{align}
Thus, $L_c$ is the integral of the norm:
\begin{equation}
L_c = \int_0^L \sqrt{ \left( 1 - a_0\left(1-\frac{s}{L}\right)\kappa(s) \right)^2 + \left(\frac{a_0}{L}\right)^2 } ds
\label{eq:L_c1}
\end{equation}
Considering the integrand as a function of $\kappa(s)$ and assuming that the local curvature due to deformation is sufficiently small, we perform a first-order Maclaurin expansion around $\kappa(s)=0$. By substituting this into Eq.~(\ref{eq:L_c1}) and rearranging the terms, we obtain the following expression:
\begin{equation}
L_c \approx \sqrt{L^2+a_0^2} - \frac{a_0 L}{\sqrt{L^2+a_0^2}} \int_0^L \left(1-\frac{s}{L}\right)\kappa(s) ds
\end{equation}
Applying integration by parts to the integral term with $\kappa(s) = d\theta/ds$ and $\theta(0)=0$:
\begin{equation}
L_c \approx \sqrt{L^2+a_0^2} - \frac{a_0}{\sqrt{L^2+a_0^2}} \int_0^L \theta(s) ds
\label{eq:L_c}
\end{equation}
Since $L, L_c$, and $a_0$ are constants, the integral $\int_0^L \theta(s) ds$ must remain constant regardless of deformation. We define the average angle $\bar{\theta}$ as a new system-specific invariant:
\begin{equation}
\bar{\theta} = \frac{1}{L} \int_0^L \theta(s) ds = \text{const.}
\end{equation}
Therefore, the length constraint of the convergent rod functions as a geometric constraint that maintains the overall average angle $\bar{\theta}$ constant.

\subsection{Mechanism of Distal Posture Preservation}
Although the distal tip angle is strictly preserved, the tip position exhibits a small variation due to geometric shortening during deformation. This occurs because bending reduces the projection of the curve along its initial direction, causing a slight deviation of the tip position.

This positional variation is not arbitrary but constrained by the invariant associated with the average angle. In particular, the tip displacement is restricted to a straight line with a constant direction determined by $\bar{\theta}$.

Based on this invariant condition, we derive how the tip coordinates are geometrically constrained. The tangent angle $\theta(s)$ after deformation can be decomposed into the constant component $\bar{\theta}$ and a local fluctuation $\delta(s)$
\begin{equation}
\theta(s) = \bar{\theta} + \delta(s) \label{eq:theta_decompose}
\end{equation}
Integrating both sides of Eq.~(\ref{eq:theta_decompose}) over $[0, L]$ yields the following property of $\delta(s)$:
\begin{equation}
\int_0^L \delta(s) ds = \int_0^L (\theta(s) - \bar{\theta}) ds = L\bar{\theta} - L\bar{\theta} = 0 \label{eq:delta_integral}
\end{equation}
The tip coordinates $x(L)$ and $y(L)$ of $\bm{r}_1(s)$ are expanded using the addition theorem:
\begin{align}
x(L) &= \int_0^L \cos(\bar{\theta} + \delta(s)) ds \nonumber \\
     &= \cos\bar{\theta} \int_0^L \cos\delta(s) ds - \sin\bar{\theta} \int_0^L \sin\delta(s) ds \label{eq:xl} \\
y(L) &= \int_0^L \sin(\bar{\theta} + \delta(s)) ds \nonumber \\
     &= \sin\bar{\theta} \int_0^L \cos\delta(s) ds + \cos\bar{\theta} \int_0^L \sin\delta(s) ds \label{eq:yl}
\end{align}
Assuming the curve is smooth and $\delta(s)$ is locally small, we approximate $\sin\delta(s) \approx \delta(s)$ using a Maclaurin expansion. From Eq.~(\ref{eq:delta_integral}), the integral terms of $\sin\delta(s)$ become nearly zero:
\begin{equation}
\int_0^L \sin\delta(s) ds \approx \int_0^L \delta(s) ds = 0
\end{equation}
Applying this to Eqs.~(\ref{eq:xl}) and (\ref{eq:yl}) eliminates the second terms, simplifying the tip coordinates to:
\begin{equation}
x(L) \approx \cos\bar{\theta} \int_0^L \cos\delta(s) ds, \quad y(L) \approx \sin\bar{\theta} \int_0^L \cos\delta(s) ds
\end{equation}
Taking the ratio of these two coordinates, the integral term $\int_0^L \cos\delta(s) ds$ cancels out:
\begin{equation}
\frac{y(L)}{x(L)} \approx \frac{\sin\bar{\theta}}{\cos\bar{\theta}} = \tan\bar{\theta} = \text{const.}
\end{equation}
This result mathematically demonstrates that the length constraint of the convergent rod passively constrains the tip to a straight line passing through the base and the initial tip position, with its direction determined by the angle $\bar{\theta}$.

\section{NUMERICAL VALIDATION}

\begin{figure}[t!]
\centering
\includegraphics[width=0.95\linewidth]{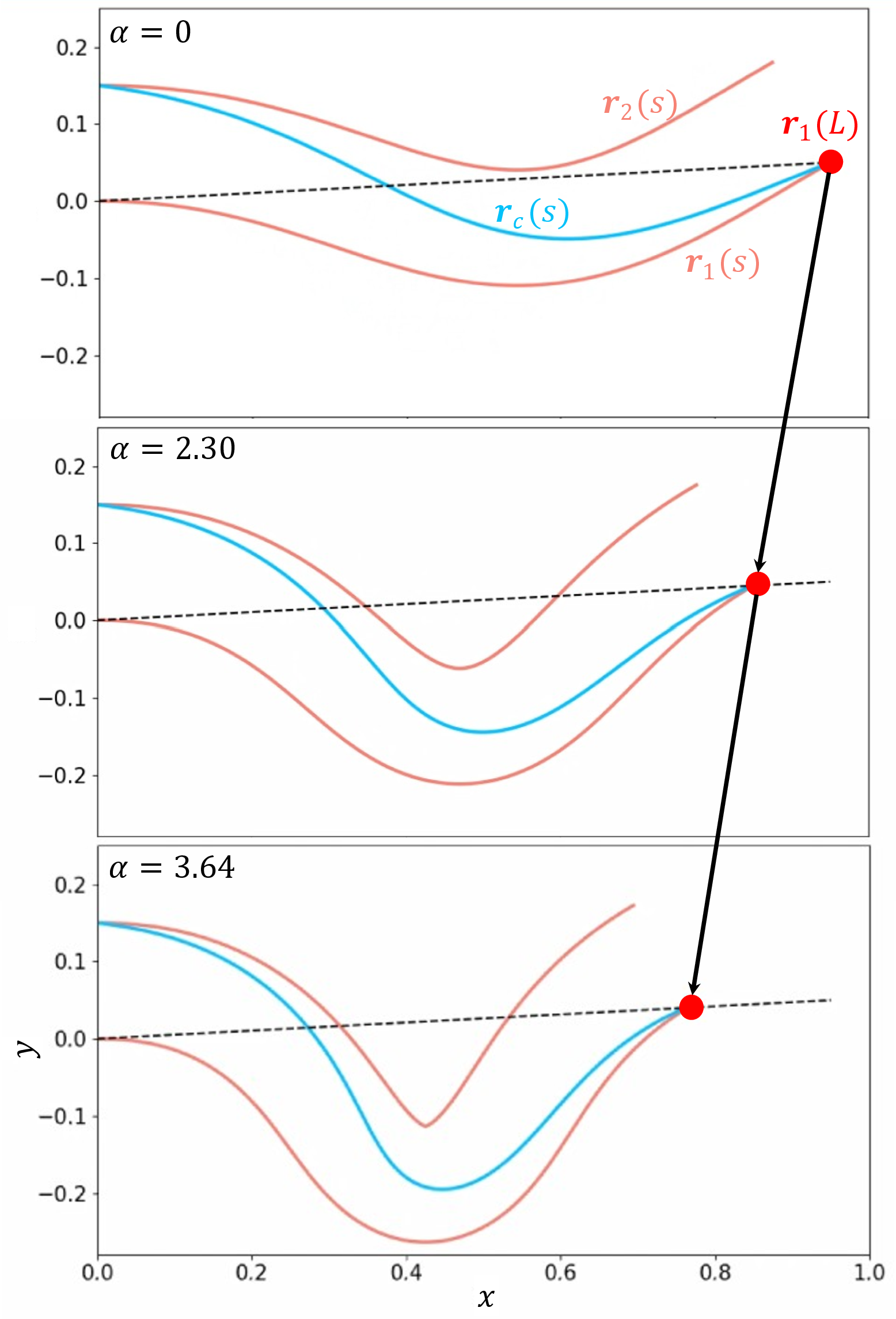}
\caption{Simulation results showing deformation while preserving distal orientation and tip trajectory.}
\label{animation}
\end{figure}

\begin{table}[t!]
    \centering
    \caption{Numerical results validating invariance of rod lengths, tip angle, and tip position ratio.}
    \begin{tabular}{cccccc}
        $\alpha$ & $L_1$ & $L_2$ & $L_c$ & $\theta(L)$ [rad] & $y(L)/x(L)$\\ \hline
        0 &   $1.00$ &   $0.92$ &   $1.01$ &        $0.52$ &       $0.053$\\
      2.30 &   $1.00$ &   $0.92$ &   $1.01$ &        $0.52$ &       $0.053$\\
      3.64 &   $1.00$ &   $0.92$ &   $1.01$ &        $0.52$ &       $0.056$\\ \hline
    \end{tabular}
    \label{tab1}
\end{table}

To visually and quantitatively verify the kinematic properties of the proposed structure (the invariance of distal orientation and tip position ratio), we developed a numerical simulation model using Python.

In this section, we first mathematically represent an arbitrary smooth deformation state by expanding the curvature distribution of the curve into a Fourier series. Next, we formulate the constraint that the length of each curve remains constant as a system of matrix equations and utilize Singular Value Decomposition (SVD) to calculate deformation vectors that alter only the shape of the curve without changing any curve lengths. Finally, we continuously apply large deformations to the curve along these vectors to quantitatively confirm that distal posture is preserved, regardless of the intermediate curvature.

\subsection{Curvature Parameterization via Fourier Series}
The curvature $\kappa(s)$ of the reference curve $\bm{r}_1(s)$ is defined as a Fourier series using basis functions $\bm{B}(s)$ to represent any continuous deformation:
\begin{equation}
\kappa(s) = \sum_{n=1}^{M} \left( a_n \cos\frac{n \pi s}{L} + b_n \sin\frac{n \pi s}{L} \right) = \bm{C}^T \bm{B}(s)
\end{equation}
where $\bm{C} \in \mathbb{R}^{2M}$ is the coefficient vector. In this simulation, we set the number of terms to $M=3$ and calculated the initial curvature $\kappa_0(s)$ to satisfy specified boundary conditions using an optimization algorithm.

\subsection{Length-Constrained Deformation Model}
The condition for constant rod lengths is expressed as two integral constraints:
\begin{enumerate}
    \item $\int_0^L \kappa(s) ds = \text{const.}$ (Invariance of distal angle)
    \item $\int_0^L s \kappa(s) ds = \text{const.}$ (Invariance of the first moment, related to average angle)
\end{enumerate}

Let the fluctuation of curvature during deformation be $\delta\kappa(s) = \bm{c}^T \bm{B}(s)$. To maintain the length constraints, the following must hold for the fluctuation:
\begin{equation}
\int_0^L \delta\kappa(s) ds = \sum_{i=1}^{2M} c_i \int_0^L B_i(s) ds = 0 
\end{equation}
\begin{equation}
\quad \int_0^L s \delta\kappa(s) ds = \sum_{i=1}^{2M} c_i \int_0^L s B_i(s) ds = 0
\end{equation}
This can be written in matrix form as $\bm{A} \bm{c} = \bm{0}$, where the elements of $\bm{A} \in \mathbb{R}^{2 \times 2M}$ consist of the integrals of the basis functions. A non-trivial coefficient vector $\bm{c}_0$ satisfying this can be calculated as the nullspace of $\bm{A}$ using SVD. This $\bm{c}_0$ represents the length-preserving deformation vector. By continuously updating the curvature as $\kappa(s) = \kappa_0(s) + \alpha \bm{c}_0^T \bm{B}(s)$ with a deformation parameter $\alpha$, We generated deformation animations that approximately satisfy the geometric constraints.

\subsection{Validation Results}
Fig.~\ref{animation} and Table \ref{tab1} show the calculated results as the parameter $\alpha$ is varied. Despite large intermediate deformations, the lengths of each curve and the distal angle $\theta(L)$ remained approximately constant. It was also confirmed that the tip position moves along the straight line connecting the origin and the initial tip position, as indicated by the dotted line in Fig.~\ref{animation}. The minor variation in $y(L)/x(L)$ is a behavior consistent with theoretical predictions, resulting from the linear approximation (neglecting higher-order terms) in the theoretical analysis, and these results quantitatively support the kinematic properties of the proposed structure.

\section{EXPERIMENTAL VALIDATION}

\begin{figure*}[t!]
  \centering
  \includegraphics[width=0.95\textwidth]{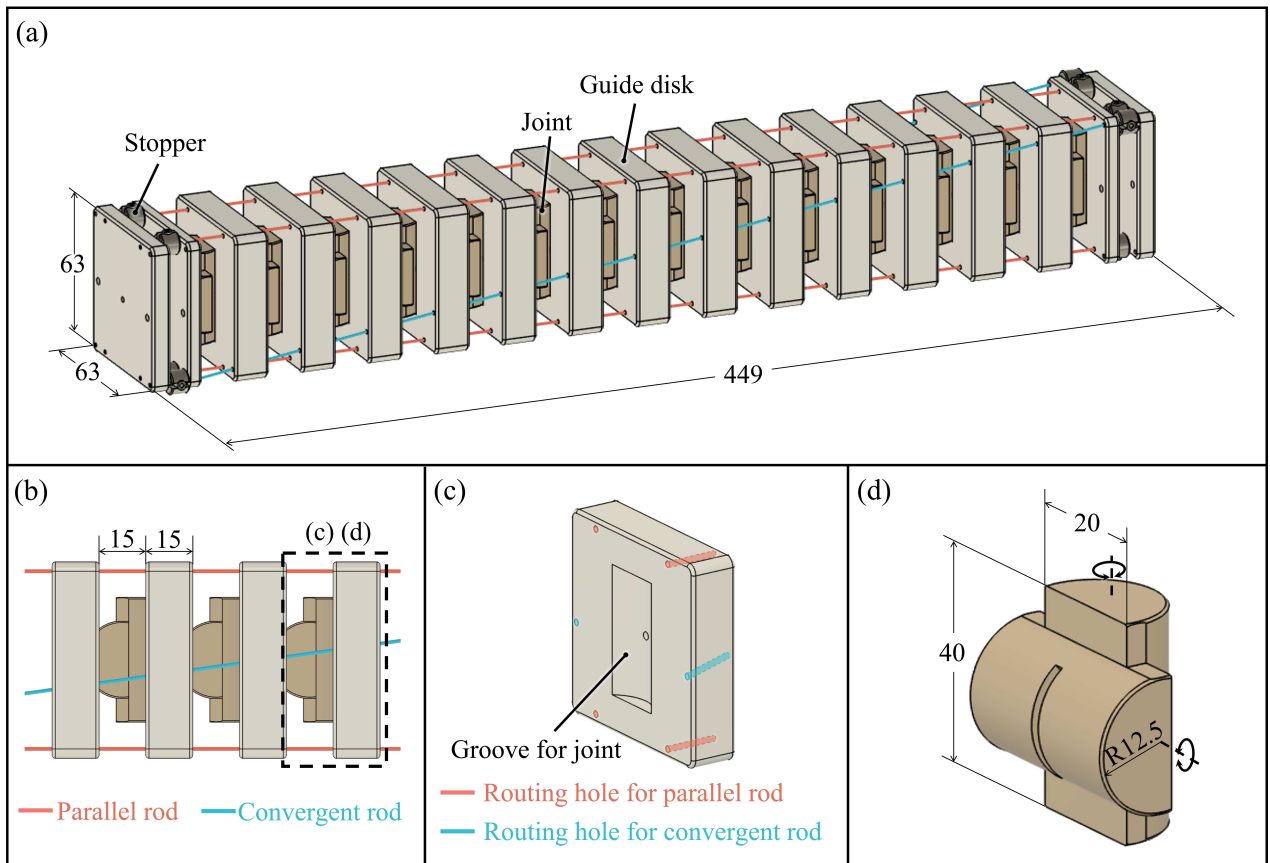}
  \caption{Design and fabrication of the prototype, including guide disks, joints, and rod configuration.}
  \label{prototype} 
\end{figure*}

\begin{figure}[t!]
\centering
\includegraphics[width=0.95\linewidth]{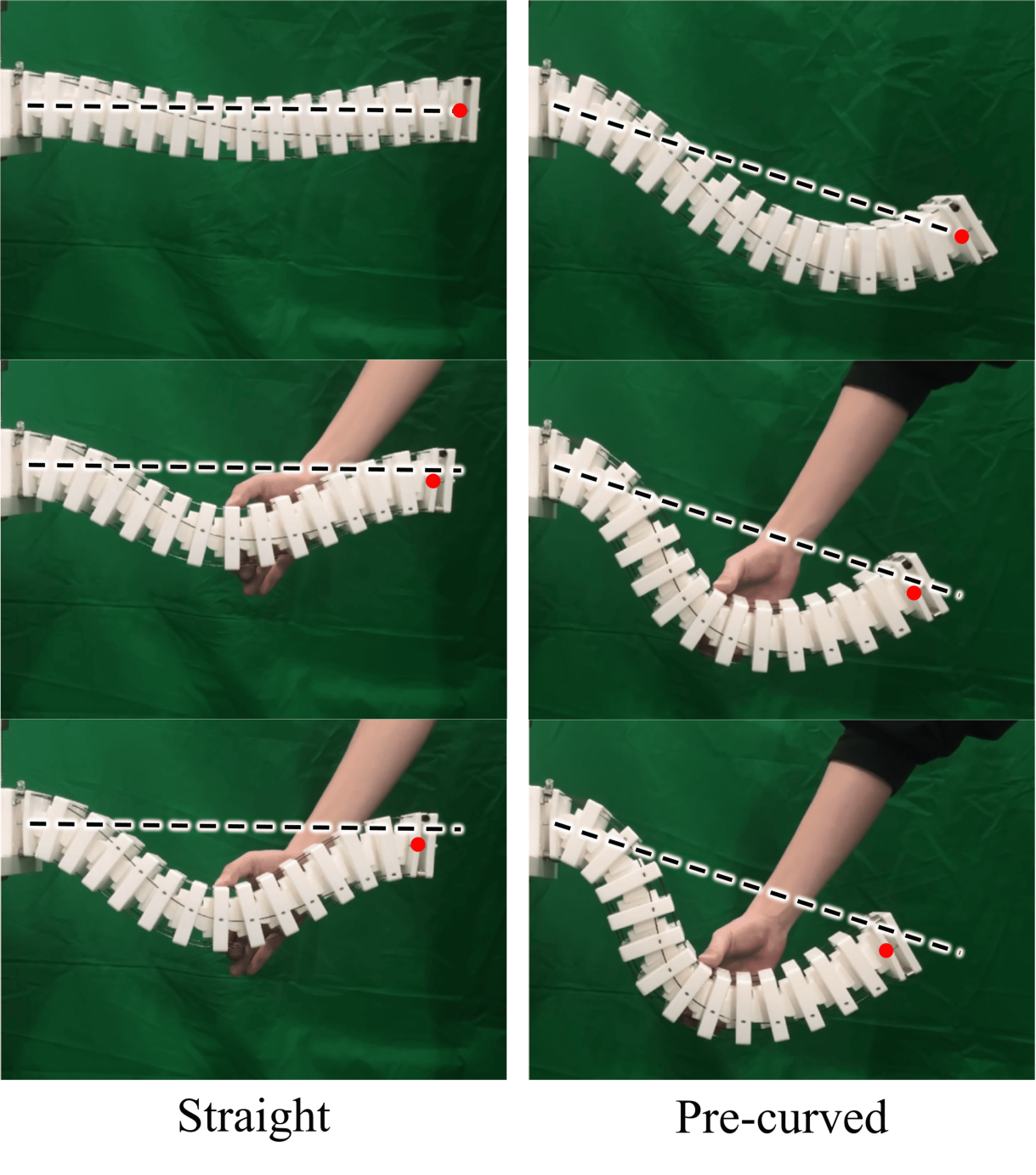}
\caption{Experimental results demonstrating that the tip follows a constrained linear trajectory under deformation.}
\label{zikken}
\end{figure}

In this section, we describe the design of the prototype and its evaluation through experiments to verify the effectiveness of the proposed structure in a physical implementation. First, we evaluate the distal behavior when the intermediate section of the prototype is forced to deform, confirming that the distal posture is passively preserved. Next, we measure the stiffness at the distal tip and the intermediate position of the prototype to demonstrate that the proposed structure achieves a high level of both intermediate compliance and distal rigidity.

\subsection{Prototype Design}
The overall configuration and side view of the designed prototype are shown in Fig.~\ref{prototype}(a) and (b). The prototype consists of guide disks and joints 3D-printed from PLA resin, and NiTi alloy rods equipped with stoppers for effective length adjustment.
The guide disks (Fig.~\ref{prototype}(c)) have a central groove for the joint and through-holes for the rods. To ensure symmetry while avoiding physical interference with the joints, a total of six NiTi rods were arranged symmetrically across the center plane. By fixing the effective length of each rod using the stoppers at the ends, the prototype can be configured into any shape.
The joint detailed in Fig.~\ref{prototype}(d) employs a 2-DOF structure consisting of two semi-cylindrical parts crossed at 90 degrees back-to-back. This joint fits into the central groove of the disks and enables smooth bending through rolling contact.

\subsection{Kinematic and Stiffness Evaluation}
In the experiments, we applied manual loads to the intermediate section of the prototype while the total lengths of the rods were fixed. Two sets of experimental conditions were established: an initially straight case and a pre-curved case. The transition of the tip coordinates during deformation was recorded using a camera, and a qualitative evaluation was conducted.

The results are shown in Fig.~\ref{zikken}. Plotting the tip coordinates revealed that in both the straight and pre-curved conditions, the tip moved along a straight line connecting the base and the initial tip position. This demonstrates that the theoretical framework holds for the physical mechanism, preserving the distal posture while allowing flexible intermediate deformation. Small deviations from the ideal trajectory are attributed to wire elastic deformation, friction, linear approximations in the theory, and discretization errors from the disks.

Furthermore, we quantitatively evaluated the stiffness at different positions by applying independent loads to the center and the distal tip. The measured stiffness was $0.18\text{ N/mm}$ at the center and $2.13\text{ N/mm}$ at the distal tip. This results in a distal-to-center stiffness ratio of approximately 12. In a conventional cantilever beam, this ratio is typically $1/8$. Thus, the proposed structure achieves a relative improvement in this ratio of approximately 100 times, confirming its ability to maintain high distal stability while keeping the intermediate section compliant.

\section{ACTUATED CONTINUUM ROBOT}
In this section, we describe the development and experimental demonstration of an actuated extensible robot arm applying the distal-stable beam principle.

\begin{figure*}[t!]
  \centering
  \includegraphics[width=0.95\textwidth]{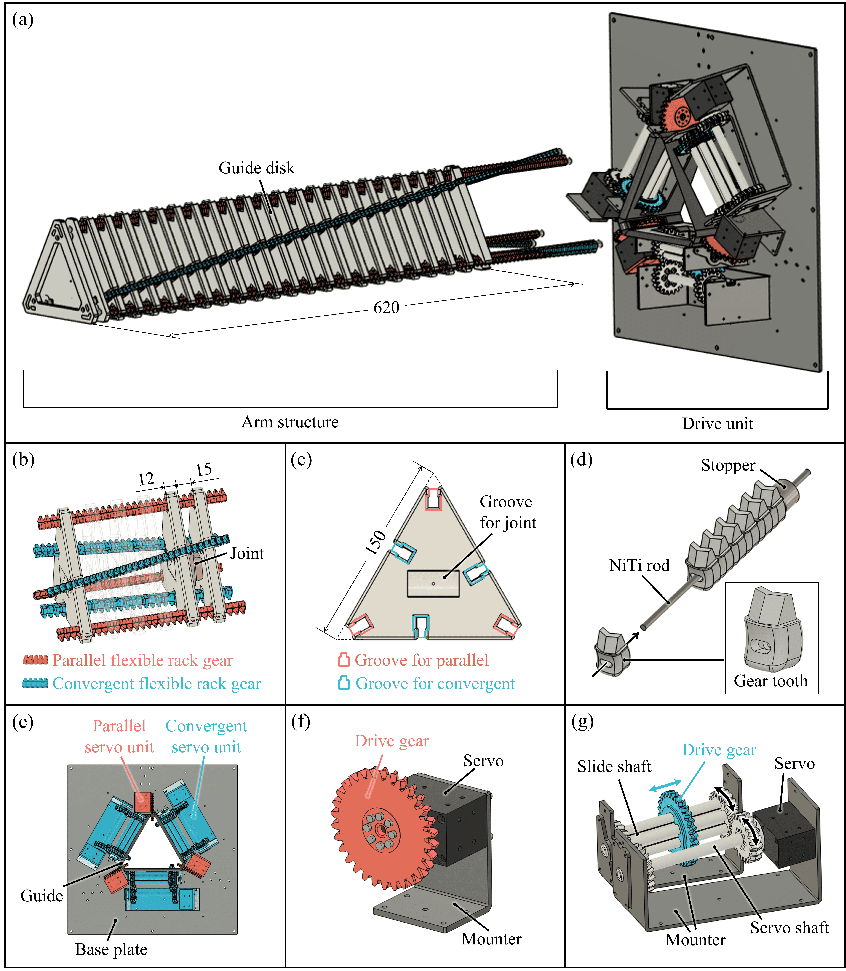}
  \caption{Design of the continuum robot incorporating the proposed distal-stable beam structure.}
  \label{robot} 
\end{figure*}

\subsection{Robot Design}
To operate the robot while maintaining the distal-stable beam structure, we adopted an extension mechanism inspired by a woodpecker's tongue~\cite{10313930}. This mechanism achieves large extension by sliding the entire arm axially. The developed robot (Fig.~\ref{robot}(a)) consists of two modules: the robot arm with the proposed beam structure and the base drive unit. The arm is composed of guide disks, rolling contact joints, and six flexible rack gears (three parallel, three convergent) for power transmission, all fabricated using 3D printing with PLA resin. The drive unit realizes arm extension and bending by sliding each rack gear individually.

The cross-sectional structure of the arm is shown in Fig.~\ref{robot}(b) and (c). The guide disks are based on an equilateral triangle, with a groove for the joint in the center, grooves for parallel rack gears near the vertices, and grooves for convergent rack gears along each side. The rack gears passing through these grooves have a NiTi rod core and multiple connected gear teeth parts (Fig.~\ref{robot}(d)). A saddle-joint shape was adopted for the contact surfaces of each tooth to suppress relative rotation around the central axis. The drive unit (Fig.~\ref{robot}(e)) consists of a guide and parallel/convergent servo units. The guide supports the arm near the base, and in the servo units (Fig.~\ref{robot}(f), (g)), the driving gears mesh with and slide the rack gears to actuate the arm. Notably, the convergent servo unit includes a mechanism allowing the driving gear to translate following the position of the convergent rack gear.

\subsection{Demonstration}

\begin{figure*}[t!]
\centering
\includegraphics[width=0.95\textwidth]{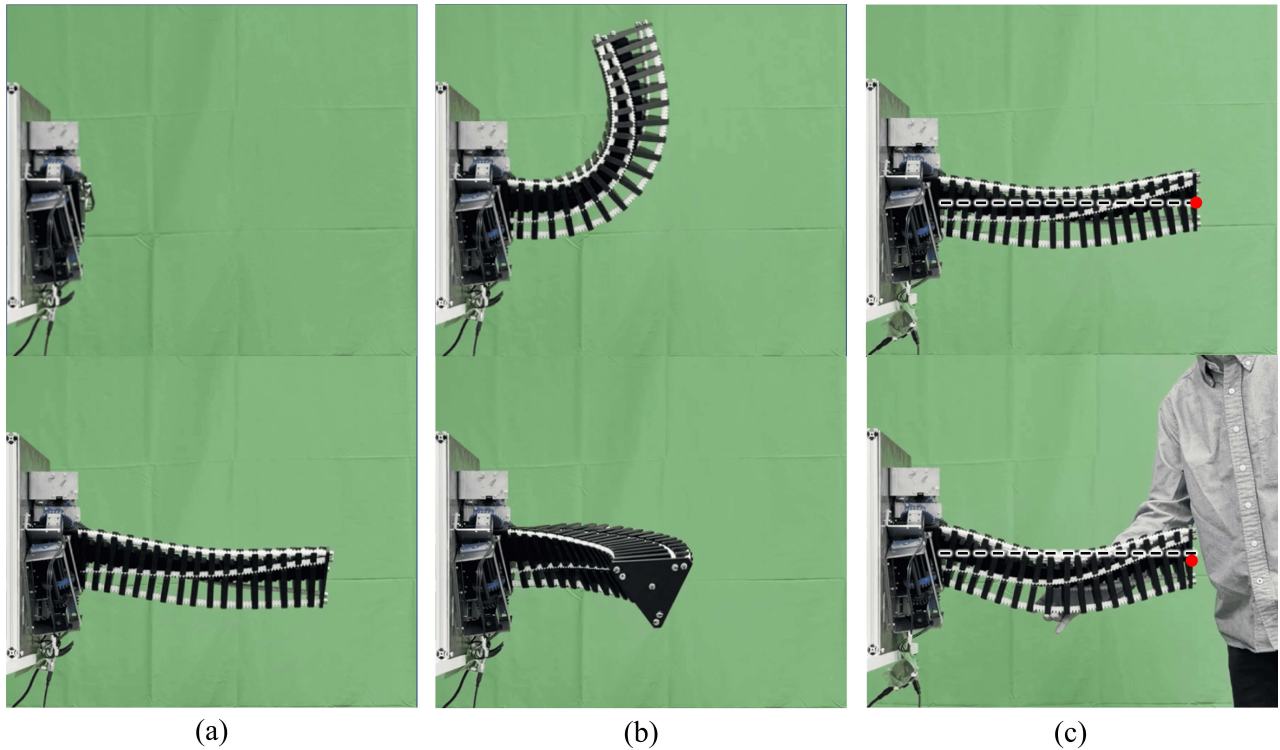}
\caption{Experimental demonstration of (a) extension, (b) bending, and (c) distal posture preservation in the robot.}
\label{demo}
\end{figure*}

To evaluate the performance of the fabricated robot, we conducted two main demonstrations.

In the first validation, actuator control demonstrated that basic continuum robot performance—extension (Fig.~\ref{demo}(a)) and 2D bending (Fig.~\ref{demo}(b))—could be performed smoothly. This proves that active deformation is possible while freely changing the total length using the flexible rack gear system.

In the second validation, we fixed the driving gears to establish the length-preserving constraint and evaluated the distal posture preservation capability against external forces applied to the intermediate section. The results showed that even with flexible intermediate deformation, the tip position and orientation remained almost invariant on the straight line, confirming the unique properties of the distal-stable beam in the actuated system (Fig.~\ref{demo}(c)).

\section{CONCLUSIONS}
In this paper, we proposed a novel structural primitive, the Distal-Stable Beam, which achieves simultaneous compliance and distal rigidity through purely geometric constraints. By introducing a triad of two parallel rods and one convergent rod, we derived invariant properties that preserve distal orientation and constrain the tip trajectory, independent of intermediate deformation. Both numerical simulations and physical experiments validated that the proposed structure maintains distal posture under large deformation while exhibiting a strong stiffness gradient, with distal stiffness significantly higher than that of the intermediate section. Furthermore, we demonstrated that this principle can be extended to an actuated continuum robot, enabling both extensibility and steerability while preserving distal stability.

The proposed mechanism represents a new approach to resolving the long-standing trade-off between softness and rigidity in continuum robotics. Unlike conventional methods relying on active stiffness modulation, the Distal-Stable Beam achieves this property passively through geometry, suggesting a new design paradigm for safe and precise robotic systems. Although demonstrated here as a robotic arm, the mechanism itself is fundamentally passive and can be applied as a general structural element, indicating broad potential across various applications beyond robotics.

There remain several important challenges for future work. First, the current analysis is purely geometric and does not provide predictive capability for stiffness. Developing a mechanics-based beam model grounded in material elasticity will be essential for quantitative design and optimization. Second, while similar rod-based continuum structures have been explored in prior work~\cite{10.3389/frobt.2021.629871}, it remains unclear why the proposed configuration uniquely exhibits strong distal posture stability. We hypothesize that the number of guide disks and their torsionally rigid coupling play a critical role in suppressing undesired deformation modes. A systematic investigation of these structural parameters is necessary to fully understand the underlying mechanism and establish general design guidelines.

\bibliography{reference}
\bibliographystyle{unsrt}

\end{document}